\documentclass[10pt,twocolumn,letterpaper]{article}

\usepackage{cvpr}              
\usepackage{comment}
\usepackage{graphicx}
\usepackage{multirow}
\usepackage{tcolorbox}
\usepackage{mwe}
\usepackage{xcolor}
\usepackage{makecell}
\usepackage{float}
\usepackage{dblfloatfix}
\usepackage{caption}
\usepackage{cuted}

\definecolor{cvprblue}{rgb}{0.21,0.49,0.74}
\usepackage[pagebackref,breaklinks,colorlinks,allcolors=cvprblue]{hyperref}

\def\paperID{} 
\def\confName{CVPR}
\def\confYear{2026}

\title{GA-VLN: Geometry-Aware BEV Representation for \\
Efficient Vision-Language Navigation}

\author{Jiahao Yang$^{1,2}$, Zihan Wang$^{4}$, Xiangyang Li$^{1,2}$, Xing Zhu$^{3}$, Yujun Shen$^{3}$, \\ Yinghao Xu$^{3,5,\dagger}$, Shuqiang Jiang$^{2,1,\dagger}$\\
\small \textsuperscript{1} State Key Laboratory of AI Safety, Institute of Computing Technology, Chinese Academy of Sciences, Beijing\\  
\small \textsuperscript{2} University of Chinese Academy of Sciences, Beijing ~~\small \textsuperscript{3} Robbyant~~\small \textsuperscript{4} School of Computing, National University of Singapore \\ \small \textsuperscript{5} The Hong Kong University of Science and Technology, Hong Kong~~~~\small $\dagger$ Corresponding authors
}

\begin{document}
\maketitle

\begin{abstract}

Despite significant progress in Vision-Language Navigation (VLN), existing approaches still rely on dense RGB videos that produce excessive patch tokens and lack explicit spatial structure, resulting in substantial computational overhead and limited spatial reasoning.
To address these issues, we introduce the Geometry-Aware BEV (GA-BEV)~–~a compact, 3D-grounded feature representation that integrates both explicit and implicit geometric cues into multimodal large language model (MLLM)~–~based navigation systems.
We construct BEV spatial maps from RGB-D inputs by projecting visual features into 3D space and aggregating them into an agent-centric layout that preserves geometric consistency while reducing token redundancy.
To further enrich geometric understanding, we incorporate features from a pretrained 3D foundation model into the BEV space, injecting structural priors learned from large-scale 3D reconstruction tasks.
Together, these complementary cues~–~explicit depth-based projection and implicit learned priors~–~yield compact yet spatially expressive representations that substantially improve navigation efficiency and performance.
Experiments show that our method achieves state-of-the-art results using only navigation data, without DAgger augmentation or mixed VQA training, demonstrating the robustness and data efficiency of the proposed GA-VLN framework.
The code is available at
\href{https://github.com/jahhaoyang/GA-VLN}{https://github.com/jahhaoyang/GA-VLN}. 

\end{abstract}    
\begin{figure}[!t]
\centering
{\includegraphics[width=\linewidth]{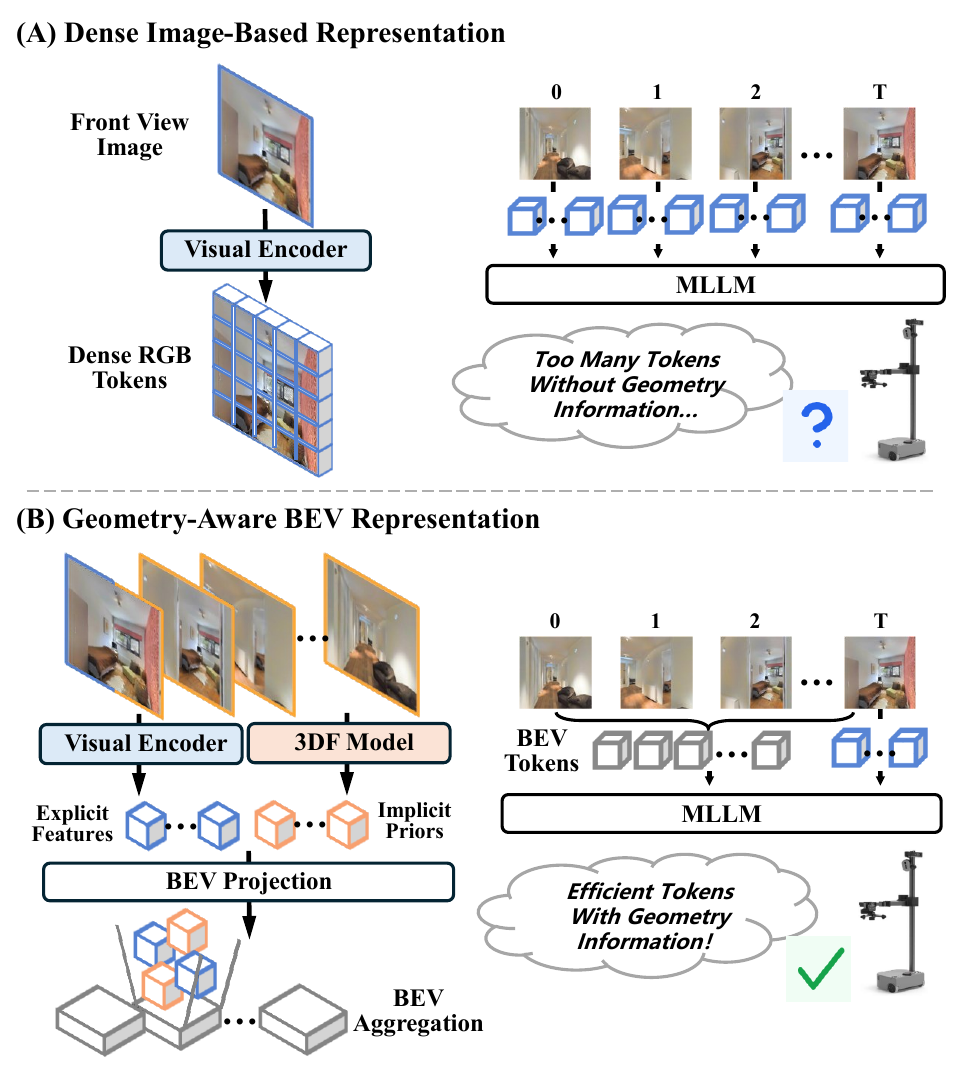}}
\caption{\textbf{Illustration of different representations for VLN}. (A) Dense image-based representations contain heavy token redundancy and lack explicit spatial structure. (B) Our Geometry-Aware BEV (GA-BEV) representation combines explicit depth-projected features with implicit geometry priors from 3D foundation models, producing a highly compact yet spatially expressive representation tailored for VLN.}
\label{fig:teaserfig}
\end{figure}

\section{Introduction}
Vision-Language Navigation (VLN)~\cite{r2r_anderson2018vision,vlnce_krantz2020beyond,Reverie_qi2020reverie,rxr_ku2020room,Navrag_wang2025navrag} aims to enable intelligent embodied agents to follow natural-language instructions and navigate through visually complex environments.
Recent advances in multimodal large language models (MLLMs)~\cite{llavaonevision_li2024llava,llavavideo_zhang2024video} have greatly enhanced agents’ abilities to comprehend instructions, ground them in visual contexts, and predict coherent action sequences.

Most existing MLLM-based VLN frameworks~\cite{Navid_zhang2024navid,Uninavid_zhang2024uni,Navila_cheng2024navila,streamvln_wei2025streamvln,auxthink_wang2025think,monodream_wang2025monodream,navfom_zhang2025embodied} process sequences of RGB frames directly as visual tokens.
While effective to some extent, this image-centric paradigm lacks explicit spatial structure and treats visual observations as flat patch embeddings without modeling geometric relationships across frames.
This results in large numbers of redundant tokens and weak spatial consistency under varying viewpoints, ultimately increasing computational cost and limiting navigation efficiency.

To address these challenges, we propose the Geometry-Aware BEV (GA-BEV)~–~a compact and spatially grounded feature representation that integrates explicit and implicit geometric cues into vision-language models for efficient VLN.
For explicit geometry awareness, we construct an agent-centric BEV map by lifting RGB-D observations from historical frames into 3D and reprojecting them onto the ground plane. This enforces spatial alignment across time and compresses redundant perspective patches into a compact, physically grounded top-down structure.
For implicit geometry awareness, we fuse geometry features from a pretrained 3D foundation model, whose large-scale 3D pretraining provides strong shape priors and structural continuity. These cues enhance the BEV map with global geometric regularities, improving robustness when depth is sparse or noisy.
Together, these complementary cues yield a BEV representation that is both compact and geometrically expressive, enabling reliable spatial reasoning for VLN.


With the proposed GA-BEV representation, we develop the geometry-aware vision-language navigation (GA-VLN) framework.
Given RGB-D inputs, patch-level visual features are first projected into 3D coordinates aligned with the agent’s pose, discretized into BEV grid cells, and aggregated to form sparse, spatially structured features.
In parallel, features from a 3D foundation model are projected into the same BEV space and fused within corresponding cells, enriching the representation with learned geometric priors.
These compact BEV features~–~together with the visual tokens from the current front-view frame and the language instruction~–~are then fed into the MLLM to predict navigation actions, enabling efficient and spatially grounded decision-making throughout the trajectory.

Extensive experiments on standard VLN benchmarks show that the proposed GA-BEV representation substantially improves navigation performance while significantly reducing the number of visual tokens for MLLM.
Furthermore, GA-VLN achieves state-of-the-art performance using only high-quality navigation data, without relying on the complex data generation of DAgger augmentation~\cite{DUET_chen2022think,Etpnav_an2024etpnav,streamvln_wei2025streamvln} or the large-scale co-training typically required for general VQA~\cite{azuma2022scanqa} datasets, which demonstrates the robustness and data efficiency of the proposed framework.

Our main contributions are summarized as follows:

\begin{itemize}
\item We propose Geometry-Aware BEV (GA-BEV), a compact and 3D-grounded representation that combines explicit depth-based projected features with implicit geometric priors from pretrained 3D foundation models.

\item We develop Geometry-Aware Vision-Language Navigation (GA-VLN), a framework that integrates GA-BEV into MLLM-based navigation, enabling improved efficiency and navigation performance.

\item Our method achieves state-of-the-art performance on standard VLN benchmarks without requiring DAgger augmentation or mixed VQA training, demonstrating the strong data efficiency of the proposed approach.

\end{itemize}




%
\section{Related Work}
\label{sec:relatedwork}   

\paragraph{Vision-Language Navigation.} The Vision-Language Navigation task~\cite{r2r_anderson2018vision,vlnce_krantz2020beyond,Reverie_qi2020reverie,rxr_ku2020room,Navrag_wang2025navrag,he2026fine} requires an embodied agent to interpret natural language instructions and navigate complex 3D environments.
Early progress focused on discrete-environment VLN~\cite{r2r_anderson2018vision,DUET_chen2022think,Gridmm_wang2023gridmm,bevbert_an2022bevbert}, where the environment is abstracted into a topological graph and the agent moves between predefined viewpoints.
In this paradigm, the agent's action space is limited to a finite set of connectivity-based choices, which restricts its ability to explore unconstrained physical spaces.
Research has since advanced toward the more challenging continuous-environment VLN~\cite{vlnce_krantz2020beyond,Etpnav_an2024etpnav}, where the agent must issue low-level control actions (e.g., move forward 0.25 m, turn left 15°) to search for the target in the continuous space.
This paradigm shift demands significantly stronger abilities in online environment representation, obstacle avoidance, and spatial memory construction, enabling the agent to understand and reason about spacial information.

\begin{figure*}[t]
    \centering
    \includegraphics[width=0.98\textwidth]{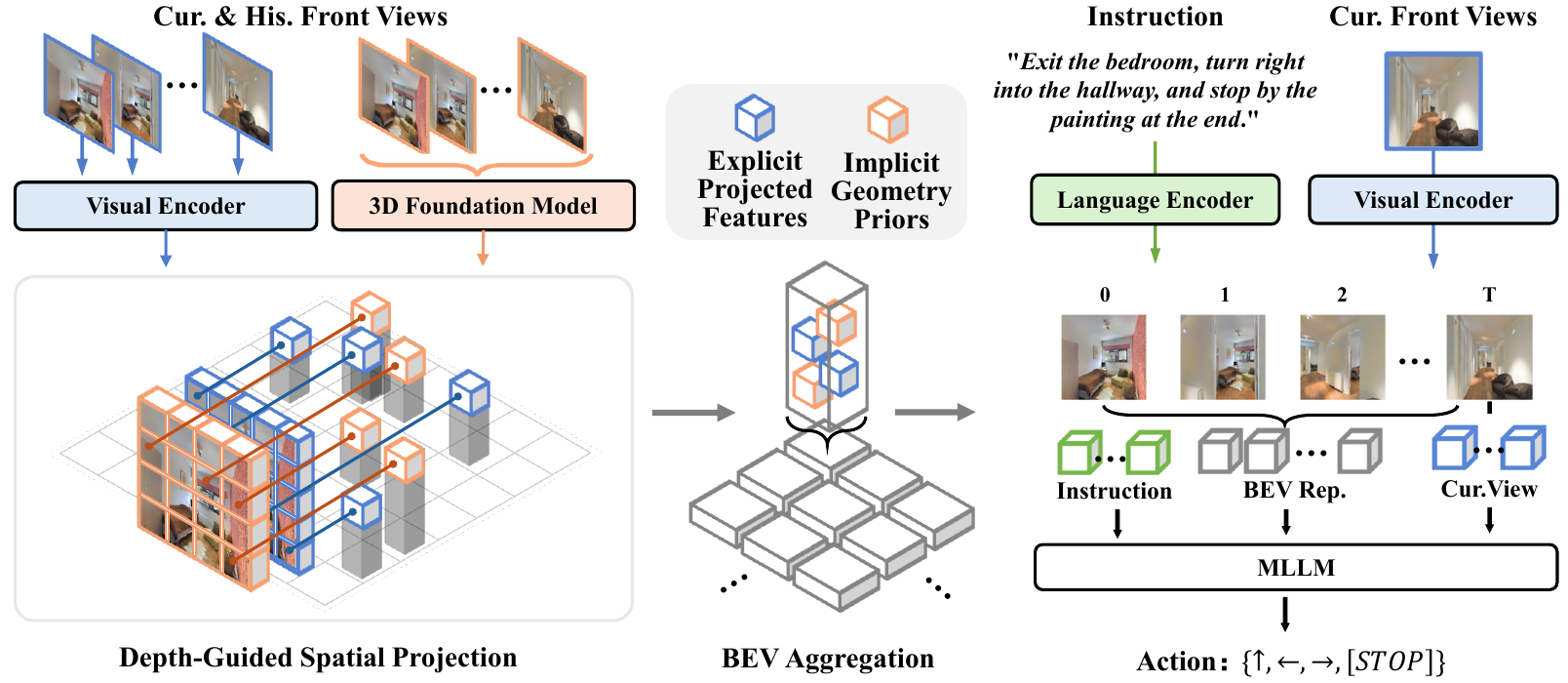}
    \caption{
        \textbf{Overview of the proposed Geometry-Aware Vision-Language Navigation (GA-VLN) framework}.
        Given RGB-D current and historical front views, our method constructs a Geometry-Aware BEV (GA-BEV) representation by combining explicit depth-guided projections with implicit geometry priors from a pretrained 3D foundation model. The projected features are aggregated into BEV grid cells to form compact and spatially expressive tokens. These BEV tokens, together with current-view features and instruction embeddings, are fed into the multimodal large language model (MLLM) to predict navigation actions.
    }
    \label{fig:placeholder}
\end{figure*}

\paragraph{Spatial Representations for VLN.}
Understanding spatial information fundamentally determines an agent’s ability to navigate in unseen environments; therefore, how to represent the surrounding environment has long been a central focus in VLN.
In discrete environments, space is abstracted into a topological map. For example, DUET~\cite{DUET_chen2022think} encodes topological nodes using a dual-scale graph transformer, fusing a coarse global plan with fine-grained local observations, and ETPNav~\cite{Etpnav_an2024etpnav} performs online topological mapping by self-organizing waypoints, effectively decoupling high-level planning from low-level control and further strengthening topological node representations.
As research progressed, several VLN works~\cite{bevbert_an2022bevbert,bevscenegraph_liu2023bird,Gridmm_wang2023gridmm} began constructing BEV maps, projecting historical observations into a top-down egocentric BEV grid.
However, due to the task formulation, these BEV maps still serve primarily as auxiliary structures for modeling spatial relations among navigation candidates, rather than as full-fledged 3D scene representations.
Pursuing higher-fidelity scene modeling, researchers further explored 3D representations methods~\cite{HNR_wang2024lookahead,Simtoreal_wang2024sim,g3dlf_wang2025g3d,lu2025monovln,Dynam3D_wang2025dynam3d} leverage principles from NeRF~\cite{mildenhall2021nerf} and 3D Gaussian Splatting~\cite{kerbl20233d} to encode or render multi-level semantic features, enabling novel-view synthesis or predictive modeling of future spatial environments.

\paragraph{MLLMs for VLN.} With the emergence and advancement of Multimodal Large Language Models (MLLMs), recent works~\cite{zheng2024towards,zhou2024navgpt,Navid_zhang2024navid,Uninavid_zhang2024uni,Navila_cheng2024navila,streamvln_wei2025streamvln} have increasingly adopted using pre-trained Multimodal Large Language Models (MLLMs) as the central policy backbone for powerful instruction comprehension. These models typically process vision as a flat sequence of RGB video frames, largely because MLLMs are trained on image-level data and therefore inherit an image-centric processing paradigm.
However, this image-centric formulation introduces two fundamental limitations: 
(1) it generates an excessive number of redundant visual tokens, incurring substantial computational overhead; and (2) it lacks explicit 3D spatial structure and multi-view geometric consistency. This deficiency limits large-scale exploration, hinders long-term environmental memory, and impairs spatial reasoning, motivating the adoption of the compact spatial representations with geometric information.

Different from all the above methods, our GA-VLN framework introduces a new paradigm by integrating a compact 3D-geometry spatial representation into an MLLM-based navigator. This approach holds distinct advantages over prior work: \textbf{1) New MLLM Paradigm.} Unlike previous BEV-based VLN models representing the navigable waypoints, GA-VLN is the first to successfully migrate the powerful instruction comprehension and reasoning of MLLMs to a spatially-aware BEV framework. \textbf{2) Computational Efficiency}. It directly addresses the bottleneck of MLLM-based navigators by replacing dense video patch tokens with our compact GA-BEV representation. This significantly reduces the number of visual tokens required, improving computational efficiency. \textbf{3) Enhanced Spatial Reasoning.} GA-VLN uniquely enriches its spatial understanding by fusing two complementary geometry sources: (1) explicit geometry from depth-guided 3D projection and (2) implicit geometric priors from a pretrained 3D foundation model. This hybrid approach provides a robust spatial context that standard image-centric pipelines lack.

\nocite{zhang3_zhang2021hierarchical,zhang4_zhang2024imagine,zhang2_zhang2022generative}

\section{Methods}
\label{sec:method}

We propose the Geometry-Aware Vision-Language Navigation (GA-VLN) framework, which incorporates a Geometry-Aware BEV (GA-BEV)~–~a compact and 3D-grounded spatial representation that transforms RGB-D observations into an agent-centric coordinate space.

In the following sections, we first review the preliminaries on existing VLN approaches. We then introduce the design details of the proposed GA-BEV representation. In Sec.~\ref{sec:action_predict}, we further describe how the GA-BEV representation is integrated into the MLLM framework for navigation.

\subsection{Preliminary}
\label{sec:preliminary}

Recent advances in vision-language models have greatly enhanced the capability of agents to reason over images and language instructions, enabling the development of powerful vision-language navigation systems in continuous environments~\cite{Navid_zhang2024navid, Uninavid_zhang2024uni, Navila_cheng2024navila,streamvln_wei2025streamvln}.
At each time step $t$, the agent receives a language instruction $L$, the current visual observation $I_t$, and all previous observations $\{ I_1, \dots, I_{t-1} \}$.
A vision-language model processes these inputs to predict the next action:
\[
a_t = f_{\text{MLLM}}\big(L, \{V_1, \dots, V_t\})
\]
where $V_i = f_v(I_i) \in \mathbb{R}^{H_p \times W_p \times d_p}$ denotes the visual tokens extracted from the RGB frame $I_i$ by the visual encoder $f_v(\cdot)$ (e.g., SigLIP~\cite{siglip_zhai2023sigmoid}),
and $f_{\text{MLLM}}(\cdot)$ represents the multi-modal large language model 
that fuses visual and linguistic inputs to infer the next action.
Existing MLLM-based pipelines~\cite{Navid_zhang2024navid,Uninavid_zhang2024uni,Navila_cheng2024navila,streamvln_wei2025streamvln}
typically feed dense patch tokens from all historical frames directly into the multimodal model, leading to substantial visual redundancy and lacking any explicit spatial structure.
This results in $t \times H_p \times W_p$ visual tokens, causing high computational overhead and weak multi-view spatial reasoning, ultimately degrading navigation performance.

\subsection{Geometry-Aware BEV Representation}
\label{sec:spatial_projection}

To achieve efficient and geometry-aware reasoning, we replace dense RGB video tokens with a compact spatial representation that explicitly encodes geometry in a BEV space.
Our proposed Geometry-Aware BEV (GA-BEV) representation provides a 3D-grounded and compact spatial encoding that integrates both \textit{explicit} and \textit{implicit} geometric cues.

\paragraph{Explicit Depth-Guided Spatial Projection.}

At each navigation step, we process the patch-level RGB image features 
$V_t \in \mathbb{R}^{H_p \times W_p \times d_p}$ and resize the corresponding 
depth map to the same resolution $D_t \in \mathbb{R}^{H_p \times W_p}$ via bicubic interpolation. 
Given the agent’s current position $\mathbf{p}_t$ and camera rotation $R_t$, 
we compute the 3D coordinates of each patch center using the pinhole camera model:

\begin{equation}
\hat{\mathbf{p}}_t(u,v)
=
\begin{bmatrix}
R_t & \mathbf{p}_t
\end{bmatrix}
K^{-1}
\begin{bmatrix}
u \\ v \\ 1
\end{bmatrix}
D_t(u,v),
\label{eq:cood}
\end{equation}
where $(u,v)$ denotes the pixel location on the image plane, 
$K$ is the camera intrinsic matrix, 
and $D_t(u,v)$ is the depth value at that location.
This back-projection maps each 2D patch center into its corresponding 3D point 
$\hat{\mathbf{p}}_t(u,v)$ in the world coordinate frame, explicitly grounding 
visual features in 3D space and injecting geometric structure at the input stage 
to enhance navigation reasoning.


\paragraph{Implicit 3D Geometry Priors.}
While the above visual representations capture explicit spatial structure from depth, they rely solely on local geometric cues within individual frames.
To incorporate broader 3D geometric priors for better spatial reasoning, we introduce representation from a pretrained 3D foundation model (e.g., VGGT~\cite{vggt_wang2025vggt}) $f_{\text{3DFM}}(\cdot)$, which encodes multi-view geometry awareness and structural regularities learned from large-scale 3D reconstruction tasks.
Specifically, we encode historical image sequences into visual patch features enriched with implicit 3D geometry priors:
\begin{equation}
V^g = f_{\text{3DFM}}\big(\{ I_1, \dots, I_t \})\in \mathbb{R}^{t\times H_g \times W_g \times d_g}
\end{equation}
Subsequently, these features are passed through a projection layer to align their dimensions with the encoded features from the visual encoder $f_v(\cdot)$: 
\begin{equation}
\tilde{V^g} = f_{\text{project}}(V^g)\in \mathbb{R}^{t\times H_g \times W_g \times d_p}
\end{equation}
We then project  $\tilde{V^g}$ into the same 3D space using the same depth-guided spatial projection procedure following Eq.~\eqref{eq:cood} to get the spatial  coordinates $\hat{\mathbf{p}^g}\in \mathbb{R}^{t\times H_g \times W_g \times 3}$. 


\paragraph{Grid-Based BEV Aggregation.}
Although the agent operates within a 3D indoor environment in VLN, its navigation trajectory is primarily constrained to the 2D ground plane. To better align with this motion constraint, we further project all 3D features onto an agent-centric BEV plane. However, features distributed in the 3D space are typically sparse , making direct aggregation inefficient. To address this, we introduce the Grid-Based BEV Aggregation method for efficient aggregation and making the representation more suitable for the navigation task.


We first define the unified feature set 
$\mathcal{V} = V \cup \tilde{V}^g$ 
and their corresponding 3D positions 
$\hat{\mathcal{P}} = \{\hat{\mathbf{p}}\} \cup \{\hat{\mathbf{p}}^{g}\}$.
All features in $\mathcal{V}$ are grounded in 3D space and projected onto the $(x, z)$ BEV plane according to their positions $\hat{\mathbf{p}}_k$.
We then aggregate features corresponding to different $y$ coordinates that fall within the same $(x, z)$ coordinate range.
Specifically, we discretize the BEV plane into a uniform $N\times N$ grid centered on the agent,
with grid cell size $\Delta$ and perception range $[-R, R]$.
For each grid cell $(i,j)$, the corresponding feature set $\mathcal{S}_{i,j}$ is defined as:
\begin{equation}
\begin{aligned}
\mathcal{S}_{i,j} = 
\big\{\, v_k \in \mathcal{V} \;\big|\;
& \hat{\mathbf{p}}_{k}^{x} \in [-R + i\Delta,\; -R + (i+1)\Delta),\\
& \hat{\mathbf{p}}_{k}^{z} \in [-R + j\Delta,\; -R + (j+1)\Delta)
\big\}.
\end{aligned}
\label{eq:bev_bin}
\end{equation}
where $\hat{\mathbf{p}}_{k} \in \mathcal{P}$, and $\hat{\mathbf{p}}_{k}^{x}$ and $\hat{\mathbf{p}}_{k}^{z}$ denote the $x$ and $z$ coordinates of the projected 3D location, respectively.

All historical 3D points $\mathcal{P}$ are transformed into the current agent-centric coordinate frame at each navigation step, ensuring that past observations remain geometrically aligned with the agent’s current pose. This reflects the inherently egocentric nature of navigation, where spatial reasoning and action decisions are made with respect to the agent’s local viewpoint.

After assigning patches to grid cells, we aggregate all features belonging to the same cell by mean pooling to form the geometry-aware BEV representation:
\begin{equation}
    B
=
\{
\frac{1}{|\mathcal{S}_{i,j}|}
\sum_{v \in \mathcal{S}_{i,j}} v
+ e_{i,j} \;\mid\; 
\begin{array}{l}
\scriptstyle |\mathcal{S}_{i,j}| > 0,\\
\scriptstyle i,j \in [1,N]
\end{array}
\},
\end{equation}
where$|\mathcal{S}_{i,j}|$ denotes the number of the features from $f_{\text{v}}(\cdot)$ and $f_{\text{3DFM}}(\cdot)$ in each grid cell, and $e_{i,j}$ is the continuous 2D sinusoidal position embedding of the agent-centric grid coordinate. And only non-empty cells are retained, meaning that the final BEV representation contains far fewer tokens than $N\times N$. It is also worth noting that although we introduce the additional $t \times H_g \times W_g$ geometry tokens from 3D foundation models, the grid-based aggregation yields a much more compact representation, resulting in even fewer total tokens than the original visual patch set $t \times H_p \times W_p$. This fusion integrates complementary geometric structures, which serve as a spatially expressive and computation-efficient representation for navigation.

\subsection{Geometry-Aware VLN Framework}
\label{sec:action_predict}

By integrating the proposed GA-BEV representation, we formulate the navigation task as an efficient two-round dialogue generation process~\cite{streamvln_wei2025streamvln} to predict discrete actions ($\mathcal{A} = \{\textit{↑, ←, →,}\texttt{STOP}\}$). As illustrated in the structured prompt below, the MLLM generates four actions per dialogue round. In the first round, the model is conditioned on the language instruction, the current front-view image, and a unified BEV feature aggregated from up to eight historical observations. To minimize computational overhead, the second round queries the model using only the newly updated front-view image while reusing the BEV feature from the first round. Consequently, the BEV representation is efficiently updated only once every eight actions (i.e., upon the completion of a two-round dialogue), and the agent terminates navigation once the \texttt{STOP} token is predicted.

\begin{tcolorbox}[
  colback=gray!10,
  colframe=black!40,
  boxrule=0.4pt,
  arc=3pt,
  left=4pt,
  right=4pt,
  top=4pt,
  bottom=4pt
]
\textbf{User:} \textit{You are an autonomous navigation assistant. Your task is to $<$INSTRUCTION$>$. 
Devise an action sequence using the actions turn left, turn right, move forward, and stop. 
\textbf{These are your bird's-eye-view feature maps from historical observations:} $<$BEV$>$. 
Your current observation is $<$IMAGE$>$.} \\[4pt]
\textbf{Assistant:} \textit{↑ ← → ↑} \\[4pt]
\textbf{User:} \textit{Your current observation is $<$IMAGE$>$.} \\[4pt]
\textbf{Assistant:} \textit{← → ↑ \texttt{STOP}}
\end{tcolorbox}

 

\section{Experiments}
\label{sec:exp}

\begin{table*}[ht]
\small
\tabcolsep=0.1cm
\centering
\begin{tabular}{c | c | c |cccc|cccc|cccc}
\toprule
\multirow{2}{*}{\textbf{Methods}} 
& \multirow{2}{*}{\textbf{System}} 
& \multirow{2}{*}{\textbf{DAgger}} 
& \multicolumn{4}{c|}{\textbf{R2R-CE}} 
& \multicolumn{4}{c|}{\textbf{RxR-CE}} 
& \multicolumn{4}{c}{\textbf{NavRAG-CE}} 
\\
\cmidrule{4-15}
& & 
& \textbf{NE\textdownarrow{}} & \textbf{OSR\textuparrow{}} & \textbf{SR\textuparrow{}} & \textbf{SPL\textuparrow{}}
& \textbf{NE\textdownarrow{}} & \textbf{OSR\textuparrow{}} & \textbf{SR\textuparrow{}} & \textbf{SPL\textuparrow{}}
& \textbf{NE\textdownarrow{}} & \textbf{OSR\textuparrow{}} & \textbf{SR\textuparrow{}} & \textbf{SPL\textuparrow{}}
\\
\midrule
CM2~\cite{cm2_georgakis2022cross} 
& \multirow{7}{*}{\makecell{Modular \\ Planner}} 
& - 
& 7.02 & 41.5 & 34.3 & 27.6 
& -    & -    & -    & - 
& -    & -    & -    & - 
\\
LAW~\cite{law_raychaudhuri2021language} 
& 
& - 
& 6.83  & 44.0 & 35.0 & 31.0 
& 10.90 & 21.0 & 8.0  & 8.0 
& -     & -    & -    & - 
\\
WS-MGMap~\cite{wsmgmap_chen2022weakly} 
&  
& - 
& 6.28 & 47.6 & 38.9 & 34.3 
& -    & -    & -    & - 
& -    & -    & -    & - 
\\
CA-Nav~\cite{canav_chen2025constraint} 
&  
& - 
& 7.58 & 48.0 & 25.3 & 10.8
& -    & -    & -    & - 
& -    & -    & -    & -  
\\
AO-Planner~\cite{aoplanner_chen2025affordances} 
&  
& - 
& 6.95 & 38.3 & 25.5 & 16.6 
& -    & -    & -    & - 
& -    & -    & -    & - 
\\
InstructNav~\cite{instructnav_long2024instructnav} 
&  
& - 
& 6.89 & - & 31.0 & 24.0 
& -    & -    & -    & - 
& 9.83 & 24.1 & 17.4 & 10.9
\\
DreamNav~\cite{dreamnav_wang2025dreamnav} 
&  
& - 
& 7.06 & 41.0 & 32.8 & 29.0 
& -    & -    & -    & - 
& -    & -    & -    & - 
\\
\midrule
VLN-3DFF~\cite{Simtoreal_wang2024sim}
& \multirow{4}{*}{3D E2E} 
& \checkmark
& 5.95 & 55.8 & 44.9 & 30.4 
& 8.79 & 36.7 & 25.5 & 18.1 
& -    & -    & -    & - 
\\
g3D-LF~\cite{g3dlf_wang2025g3d} 
&  
& \checkmark
& 5.70 & 59.5 & 47.2 & 34.6 
& -    & -    & -    & - 
& 8.85 & 31.8 & 21.4 & 13.5
\\
MapNav~\cite{mapnav_zhang2025novel} 
&  
& \checkmark
& 4.93 & 53.0 & 39.7 & 37.2 
& 8.95 & -    & 22.1 & 20.2 
& -    & -    & -    & - 
\\
Dynam3D~\cite{Dynam3D_wang2025dynam3d} 
&  
& \checkmark
& 5.34 & 62.1 & 52.9 & 45.7 
& -    & -    & -    & - 
& 8.12 & 38.4 & \textbf{24.7} & \textbf{18.8}
\\
\midrule
NaVid~\cite{Navid_zhang2024navid} 
& \multirow{7}{*}{\makecell{Image-based \\ MLLM}} 
& \checkmark 
& 5.47 & 49.1 & 37.4 & 35.9 
& -    & -    & -    & - 
& 9.35 & 29.6 & 19.4 & 13.9
\\
Uni-NaVid~\cite{Uninavid_zhang2024uni} 
&  
& \checkmark
& 5.58 & 53.3 & 47.0 & 42.7 
& 6.24 & 55.5 & 48.7 & 40.9 
& -    & -    & -    & - 
\\
NaVILA~\cite{Navila_cheng2024navila} 
&  
& $\times$
& 5.22 & 62.5 & 54.0 & 49.0 
& 6.77 & -    & 49.3 & 44.0 
& -    & -    & -    & - 
\\
Aux-Think~\cite{auxthink_wang2025think} 
&  
& \checkmark
& 6.08 & 60.0 & 54.8 & 46.9 
& 6.24 & 61.9 & 52.2 & 40.2 
& -    & -    & -    & - 
\\   
MonoDream~\cite{monodream_wang2025monodream}  
&  
& \checkmark 
& 5.45 & 61.5 & 55.8 & 49.1 
& -    & -    & -    & - 
& -    & -    & -    & - 
\\
StreamVLN~\cite{streamvln_wei2025streamvln} 
&  
& \checkmark
& 4.98 & 64.2 & 56.9 & 51.9 
& 6.22 & -    & 52.9 & 46.0
& -    & -    & -    & - 
\\
InternVLA-N1~\cite{internnav2025} 
&  
& \checkmark
& 4.83 & 63.3 & 58.2 & 54.0 
& 5.91 & - & 53.5 & \textbf{46.1}  
& - & - & - & - 
\\
\midrule
\textcolor{black}{Ours} 
& GA-VLN
& $\times$
& \textbf{4.80} & \textbf{67.6} & \textbf{61.0} & \textbf{55.2} 
& \textbf{5.88} & \textbf{67.0} & \textbf{55.4} & 45.2
& \textbf{7.88} & \textbf{46.4} & 22.2 & 18.2
\\
\bottomrule 
\end{tabular}
\vspace{-3pt}
\caption{\textbf{Comparison with state-of-the-art VLN methods on R2R-CE, RxR-CE, and NavRAG-CE val$\_$unseen benchmarks}. ``System'' groups methods into modular planners, 3D end-to-end agents, and Image-based MLLM agents, while ``DAgger'' indicates the use of DAgger augmentation data.}
\label{tab:navigation_sota}
\end{table*}

\subsection{Experimental Setup}

\paragraph{Benchmarks and Metrics.}
We evaluate our approach on standard continuous-environment VLN-CE~\cite{vlnce_krantz2020beyond} benchmarks: R2R-CE~\cite{r2r_anderson2018vision}, RxR-CE~\cite{rxr_ku2020room}, and NavRAG-CE~\cite{Navrag_wang2025navrag} val\_unseen split in the Habitat simulator~\cite{Habitat_savva2019habitat}. 
We adopt the monocular vision-and-language navigation in continuous-environments setting, where the agent’s observations are limited to a 60-degree field-of-view image directly facing forward during the navigation process. 
Navigation performance is measured using four standard metrics: Navigation Error (NE), Success Rate (SR), Oracle Success Rate (OSR), and Success weighted by Path Length (SPL).

\paragraph{Training Data.}
Our model is trained on a combination of navigation datasets collected in MP3D~\cite{Matterport3d_chang2017matterport3d} and HM3D~\cite{hm3d_ramakrishnan2021habitat} environments, including:
R2R-CE~\cite{r2r_anderson2018vision} (10,819 trajectories), 
RxR-CE~\cite{rxr_ku2020room} (19,990 trajectories), 
EnvDrop~\cite{envdrop_tan2019learning} (146,304 trajectories), 
ScaleVLN~\cite{ScaleVLN_wang2023scaling} (155,098 trajectories), 
and SRDF~\cite{SRDF_wang2024bootstrapping} (319,022 trajectories). 
The SRDF dataset is constructed by transferring SRDF trajectories into the continuous VLN setting and filtering to retain high-quality data.
All other datasets follow settings consistent with ~\cite{streamvln_wei2025streamvln}.
No DAgger~\cite{DUET_chen2022think,Etpnav_an2024etpnav,streamvln_wei2025streamvln} augmentation data or general VQA~\cite{azuma2022scanqa} datasets are used in our work.

\paragraph{Implementation Details.}
We adopt LLaVA-Video-7B~\cite{llavavideo_zhang2024video} as the base $f_{\text{MLLM}}$ with the visual encode $f_{\text{v}}$ SigLIP~\cite{siglip_zhai2023sigmoid}.
For BEV representation settings, grid cell size $\Delta$ is 0.25 meters, BEV range is $[$-10 meters, 10 meters$]$.
For the 3D foundation model $f_{\text{3DFM}}$, we use VGGT-1B~\cite{vggt_wang2025vggt} and extract features from its penultimate layer~\cite{vgllm_zheng2025learning}, followed by $f_{\text{project}}$~–~a 2-layer MLP (Linear–GeLU–Linear) with a 4096-dimensional hidden layer to match the SigLIP embedding dimension.
The parameters of $f_{\text{3DFM}}$ are kept frozen during training, while all other modules are fine-tuned. The model is optimized using a cosine annealing schedule with a minimum learning rate. We set the learning rate of the visual encoder to 5e-6 and that of all other components to 2e-5. All reported results are obtained using models pretrained for 2 epochs. 

\subsection{Comparison with State-of-the-Art Methods}
\label{sec:sota}

We compare our approach with a comprehensive set of state-of-the-art monocular VLN methods in continuous environments, including modular planners, 3D end-to-end agents, and recent image-based MLLM frameworks. Table~\ref{tab:navigation_sota} reports results on the R2R-CE, RxR-CE, and NavRAG-CE benchmarks. Across most metrics on these benchmarks, our GA-VLN achieves the best overall performance, consistently surpassing previous Image-based MLLM frameworks~\cite{Navid_zhang2024navid, Uninavid_zhang2024uni, Navila_cheng2024navila, streamvln_wei2025streamvln} and pipelines~\cite{Dynam3D_wang2025dynam3d,g3dlf_wang2025g3d,Simtoreal_wang2024sim} that explicitly utilized depth information. These improvements showcase the strong inductive bias introduced by our GA-BEV representation, which provides spatially structured geometric context and facilitates efficient multimodal fusion for navigation. 
To avoid cross-dataset interference caused by the significant distribution gap, we do not incorporate the NavRAG-CE training data into our main training corpus.
The NavRAG-CE results are obtained by finetuning the final model for one additional epoch on the NavRAG-CE training set.

\begin{table*}[t]
  \centering
  \tabcolsep=0.1cm
  \small
  \begin{tabular}{cl | c | c | cccc | ccc | c}
    \toprule
    \multirow{2}{*}{} & \multirow{2}{*}{\textbf{Method}} & \multirow{2}{*}{\textbf{BEV Rep.}} & \multirow{2}{*}{\textbf{3D-Geo.}} & \multicolumn{4}{c|}{\textbf{R2R-CE Performance}} & \multicolumn{3}{c|}{\textbf{Avg. TFLOPs diff.}} & \textbf{Lat.} \\
    \cmidrule(lr){5-8} \cmidrule(lr){9-11}
    & & & & \textbf{NE}\textdownarrow{} & \textbf{OSR}\textuparrow{} & \textbf{SR}\textuparrow{} & \textbf{SPL}\textuparrow{} & \textbf{MLLM} & \textbf{Others} & \textbf{Total} & \textbf{(ms/Inf)} \\
    \midrule
    \#1 & Baseline & $\times$   & $\times$   & 5.89 & 58.83 & 51.49 & 46.18 & 32.19 & - & 32.19 & 342.9 \\
    \#2 & GA-VLN (w/o VGGT) & \checkmark & $\times$   & 4.86 & 64.82 & 59.21 & 53.87 & 5.15 & - & 5.15 & \textbf{212.9} \\
    \#3 & GA-VLN & \checkmark & \checkmark & \textbf{4.80} & \textbf{67.59} & \textbf{60.96} & \textbf{55.19} & 6.76 & 1.97 & 8.73 & 258.7 \\      
    \bottomrule  
  \end{tabular}
  \caption{Ablation study of Geometry-Aware BEV representation and efficiency comparison per inference step.}
  \label{tab:Ablation}
\end{table*}

\begin{table*}[t]
  \centering
  \small
  \setlength{\tabcolsep}{3pt}
  \begin{tabular}{cl | c c c | c | c c c c}
    \toprule
    & \textbf{Method}
    & \textbf{Bev Grid Size} 
    & \textbf{3D-Geo.}
    & \textbf{BEV Steps}
    & \textbf{Token Num} 
    & \textbf{NE}\textdownarrow{} 
    & \textbf{OSR}\textuparrow{} 
    & \textbf{SR}\textuparrow{} 
    & \textbf{SPL}\textuparrow{}
    \\
    \midrule
    \#1 & Baseline & - & $\times$ & 32
    & 4003 & 6.08 & 54.59 & 46.49 & 42.36 \\
    \#2 & GA-VLN (w/o VGGT) & 0.25m$\times$0.25m & $\times$ & 32
    & 394 & 5.33 & 56.33 & 51.50 & 48.25 \\
    \#3 & GA-VLN & 0.25m$\times$0.25m & \checkmark & 32
    & 514 & 5.03 & 59.60 & 53.56 & 49.41 \\
    \midrule
    \#4 & & 0.125m$\times$0.125m & \checkmark & 32
    & 1193 & 5.35 & 55.90 & 51.27 & 47.55 \\
    \#5 & & 0.5m$\times$0.5m  & \checkmark & 32
    & 184 & 5.38 & 57.53 & 50.52 & 46.41 \\    
    \midrule
    \#6 & & 0.25m$\times$0.25m & \checkmark & 48
    & 570 & 5.02 & 62.48 & 54.38 & 48.99 \\
    \#7 & & 0.25m$\times$0.25m & \checkmark & 64
    & 592 & 5.14 & 62.31 & 53.56 & 48.08 \\
    \#8 & & 0.25m$\times$0.25m & \checkmark & 96
    & 599  & 5.20 & 61.71 & 53.45 & 48.02 \\    
    \bottomrule 
  \end{tabular}
  \caption{Analysis of token efficiency and spatial resolution trade-offs of GA-BEV. The experiments compare different visual representations (rows 1–3), BEV grid size (rows 4–5), and BEV step range (rows 6–8). “Token Num” denotes the total visual tokens fed into the $f_{\text{MLLM}}$. Unlike Table~\ref{tab:Ablation}, all models in this table are trained without incorporating the SRDF dataset to reduce computational overhead.}
  \label{tab:tokennum}
\end{table*}


Notably, unlike most recent methods that rely on DAgger augmentation, our framework achieves competitive performance using only high‑quality curated data, without any DAgger-enhanced trajectories or large‑scale VQA co‑training. This highlights the data efficiency and strong inductive bias of the GA‑BEV representation, eliminating the need for labor‑intensive DAgger data generation while maintaining strong navigation performance.

\begin{figure*}[t]
    \centering
    \includegraphics[width=1\textwidth]{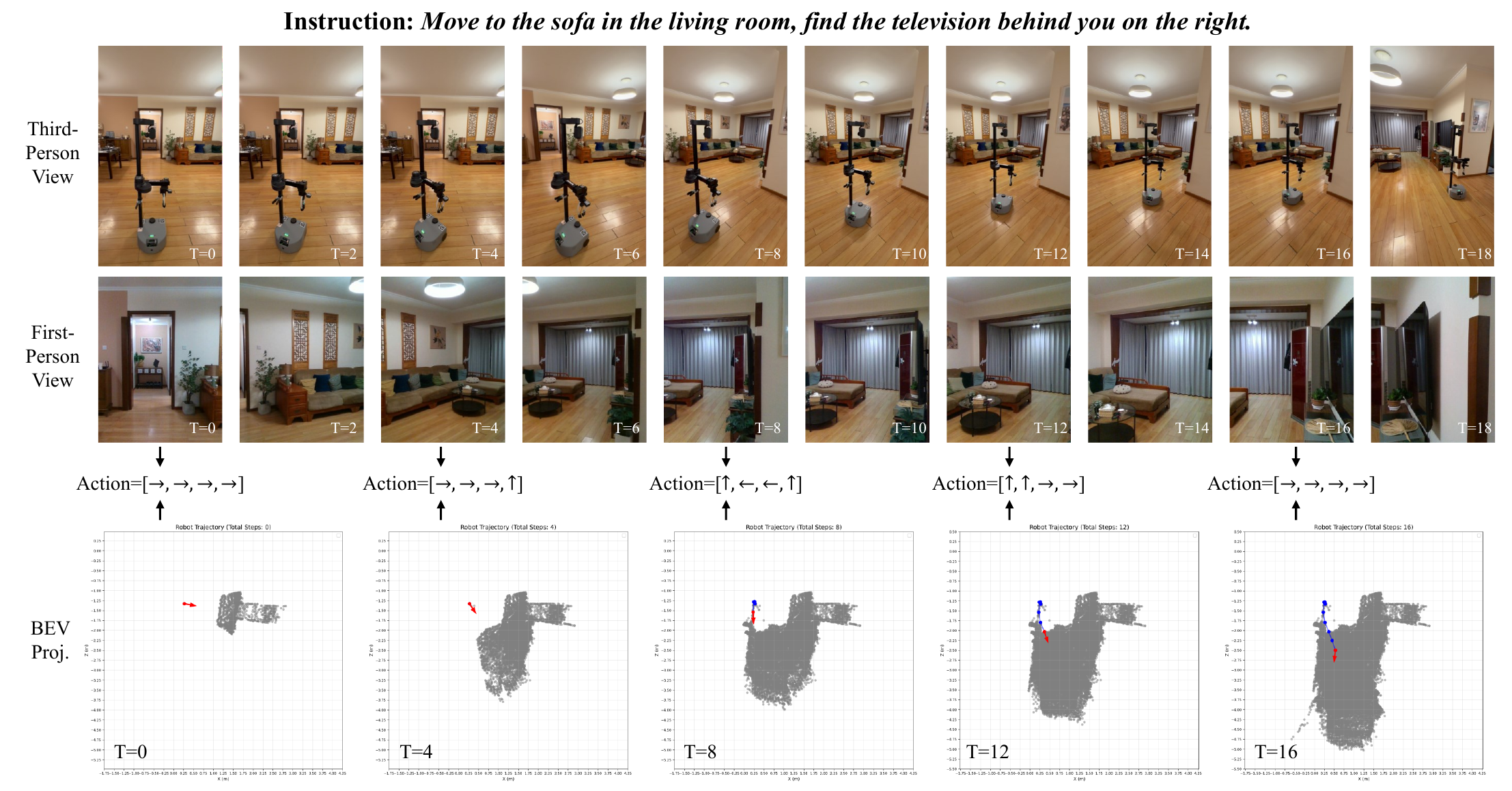}
    \captionof{figure}{An example of the GA-VLN real-world result.}
    \label{fig:exmp1}
\end{figure*}

\subsection{Ablation Study and Efficiency Analysis}
Table~\ref{tab:Ablation} presents the main ablation study of the proposed Geometry-Aware BEV representation. 
``BEV Rep.'' indicates that explicit depth-guided spatial projection is applied to map visual features into the BEV space; 
``3D-Geo.'' denotes that implicit 3D geometry priors extracted from the pretrained 3D foundation model are fused with 2D visual features. All the ablation and analysis experiments in this section and the following sections are conducted on the R2R-CE val\_unseen split.

Comparing rows \#1 vs.~\#2 demonstrates the effectiveness of explicit depth-guided BEV projection. 
By introducing explicit spatial information, the BEV features enable the agent to better capture the surrounding environment, resulting in improved spatial understanding and higher navigation accuracy.

Rows \#2 vs.~\#3 further validate the benefit of incorporating implicit 3D geometry priors from the pretrained 3D foundation model, which enhances spatial awareness by injecting multi-view geometric knowledge. 

Finally, the full GA-BEV configuration integrating ``BEV Rep.'' and ``3D-Geo.'' together achieves the best overall performance, demonstrating that explicit depth-guided BEV projection and implicit 3D geometry priors are highly complementary.
Their combination strengthens spatial reasoning, enhances data efficiency, and yields a more robust navigation representation.

Furthermore, Table \ref{tab:Ablation} reports the per-inference theoretical TFLOPs and model latency evaluated on identical samples and hardware to demonstrate the computational efficiency of our method. By elegantly compressing historical observations into our GA-BEV representation, the MLLM inference cost drops significantly. While processing frames through the VGGT-1B and the subsequent 2-layer MLP projection introduces an additional marginal overhead, the overall computational burden remains substantially lower than the baseline. Ultimately, GA-VLN outperforms the image-based baseline in both navigation performance and inference speed. This demonstrates the superior efficiency of our BEV-based representation, which accommodates rich 3D geometric priors at a strictly controlled and highly acceptable latency cost.


\subsection{Design Analysis of GA-BEV}

Table~\ref{tab:tokennum} analyze the factors that influence the efficiency of GA-BEV, such as the number of tokens and the spatial granularity in the BEV representation. The token numbers represent the average number of visual tokens required per navigation step, computed over 121 sampled navigation trajectories across 61 training scenes.

\paragraph{Effectiveness under Restricted Data Budgets.}
To rigorously demonstrate that the performance gains of our model are driven by fundamental architectural innovations rather than solely by data scaling, Rows \#1--3 in Table~\ref{tab:tokennum} evaluate the core components of GA-VLN under a restricted data budget, which was trained without the SRDF dataset. This setup directly mirrors the ablation study in Table~\ref{tab:Ablation}, allowing for a fair comparison of our method across different data scales. 
Crucially, the consistent relative improvements observed both with the SRDF dataset (Table~\ref{tab:Ablation}) and without it (Table~\ref{tab:tokennum}) confirm that GA-VLN provides a robust spatial inductive bias that is independent of the training data volume. This indicates that our architectural design and data scaling strategies act as complementary, rather than conflicting, drivers for achieving state-of-the-art performance.

\paragraph{Architectural Innovation vs. Data Scaling.} 
As shown in Table~\ref{tab:tokennum} (Rows \#1--3), the ablation trends under a restricted data budget (w/o SRDF) perfectly align with those in Table~\ref{tab:Ablation}. Specifically, compressing observations into our GA-BEV representation (Row \#2) improves the SR from 46.49\% to 51.50\% while drastically reducing visual tokens from 4003 to 394. Incorporating 3D-geometric priors via VGGT (Row \#3) further pushes the SR to 53.56\% with a manageable 514 tokens. Crucially, these consistent relative improvements across different data scales confirm that GA-VLN provides a robust spatial inductive bias independent of data volume. Thus, our architectural design and data scaling act as complementary, rather than conflicting, drivers of performance.

\paragraph{Effect of BEV Grid Size.}
Rows \#3 – \#5 in Table~\ref{tab:tokennum} evaluate the impact of BEV grid size. 
The results show that a moderate resolution (0.25\,m~$\times$~0.25\,m) achieves the best trade-off between accuracy and efficiency. 
An overly fine grid (row \#4) fails to effectively compress redundant features, while an overly coarse grid (row \#5) leads to the loss of important spatial details.

\paragraph{Effect of the Number of Historical Frames.}
Rows \#3 and \#6 – \#8 in Table~\ref{tab:tokennum} investigate the influence of the number of historical frames used to construct the BEV representation. 
While performance slightly improves when extending the temporal window from 32 to 48 action steps, it saturates or even decreases with longer histories.
This indicates that a 32-step temporal range already provides sufficient environmental context for spatial reasoning and navigation, whereas more distant observations offer limited benefit due to their reduced spatial relevance and accumulated noise in the BEV space. In addition, the action steps have only a minor influence on the token count throughout the navigation process, validating the use of a compact historical window to maintain a high-quality yet efficient BEV representation.

\subsection{Real-World Robot Experiments.} To validate the zero-shot generalizability of GA-VLN, we deploy it on a physical Hello Robot Stretch 3 in a real-world room. As shown in Fig.~\ref{fig:exmp1}, despite operating without any auxiliary obstacle avoidance or mapping modules, the agent successfully executes complex natural-language instructions and constructs meaningful geometric surrounding BEV representations. GA-VLN demonstrates strong instruction comprehension and highly reliable physical execution capabilities. Detailed hardware setups and additional examples are provided in the Supplementary Material.

\begin{table}[t]
\renewcommand{\arraystretch}{0.85}
\setlength{\tabcolsep}{6pt}
\setlength{\aboverulesep}{1pt}
\setlength{\belowrulesep}{1pt} 
\centering
\caption{Robustness to Sensor Noise on R2R-CE val unseen.}
\label{tab:test_noise}
\vspace{-7pt}
\footnotesize
\begin{tabular}{cccccc}
\toprule
\textbf{Noise} & \textbf{$\mathcal{N}(0, \sigma^2)$} & \textbf{NE$\downarrow$} & \textbf{OSR$\uparrow$} & \textbf{SR$\uparrow$} & \textbf{SPL$\uparrow$} \\ \midrule
w/o Noise & - & 4.80 & \textbf{67.59} & \textbf{60.96} & \textbf{55.19} \\ \midrule
Depth & $\sigma = 0.05\text{m}$ & 4.82 & 65.63 & 59.11 & 54.25 \\
Pose & $\sigma = 0.05\text{m}$ & 4.78 & 67.16 & 59.82 & 54.27 \\
Rotation & $\sigma = 5^{\circ}$ & 4.93 & 66.50 & 58.29 & 52.99 \\ 
\bottomrule
\end{tabular}
\end{table}

\subsection{Robustness to Noise.} Table.~\ref{tab:test_noise} evaluates GA-VLN under noise levels modeled after real-world error profiles of Stretch 3 robot. The marginal performance drop (\textless2\% SR drop) suggests that our spatial BEV aggregation and depth down-sampling as filters against depth jitter and pose drift. We posit that the cross-view consistency from VGGT contributes to a more stable geometric map, making GA-VLN more reliable than brittle visual-only cues.

\section{Conclusion}
\label{sec:conclusion}

In this work, we propose the Geometry-Aware Vision-Language Navigation (GA-VLN) framework, which introduces a Geometry-Aware BEV (GA-BEV) representation to enhance spatial reasoning in multimodal large language models.
GA-BEV explicitly grounds visual features in 3D space via depth-guided projection and integrates implicit geometric priors from a pretrained 3D foundation model.
Through grid-based aggregation, these features are organized into an agent-centric BEV map, forming a compact and spatially expressive representation for navigation.
Experiments on standard VLN benchmarks show that GA-VLN achieves state-of-the-art performance with substantially fewer visual tokens, demonstrating that coupling explicit and implicit 3D geometry greatly enhances agents’ spatial awareness and navigation efficiency.

\paragraph{Acknowledgements:}
This work was supported in part by the National Natural Science Foundation of China under
Grants 62495084, 62125207, U23B2012, 62102400, and 62272436, in part by the National Postdoctoral Program for Innovative Talents under Grant BX20200338, in part by the Suzhou Science and Technology Plan Project under grant SYG2024082.

{
    \small
    \bibliographystyle{ieeenat_fullname}
    \bibliography{main}
}

\clearpage

\setcounter{page}{1}

\maketitlesupplementary

\appendix

\section{Real-World Robot Experiments}
\label{sec:realworld}

To validate the effectiveness and generalizability of the proposed GA-VLN framework, we deploy it onto the physical robot and conduct qualitative VLN experiments in real-world environments. 

\begin{figure*}[b]
    \centering
    \includegraphics[width=1\textwidth]{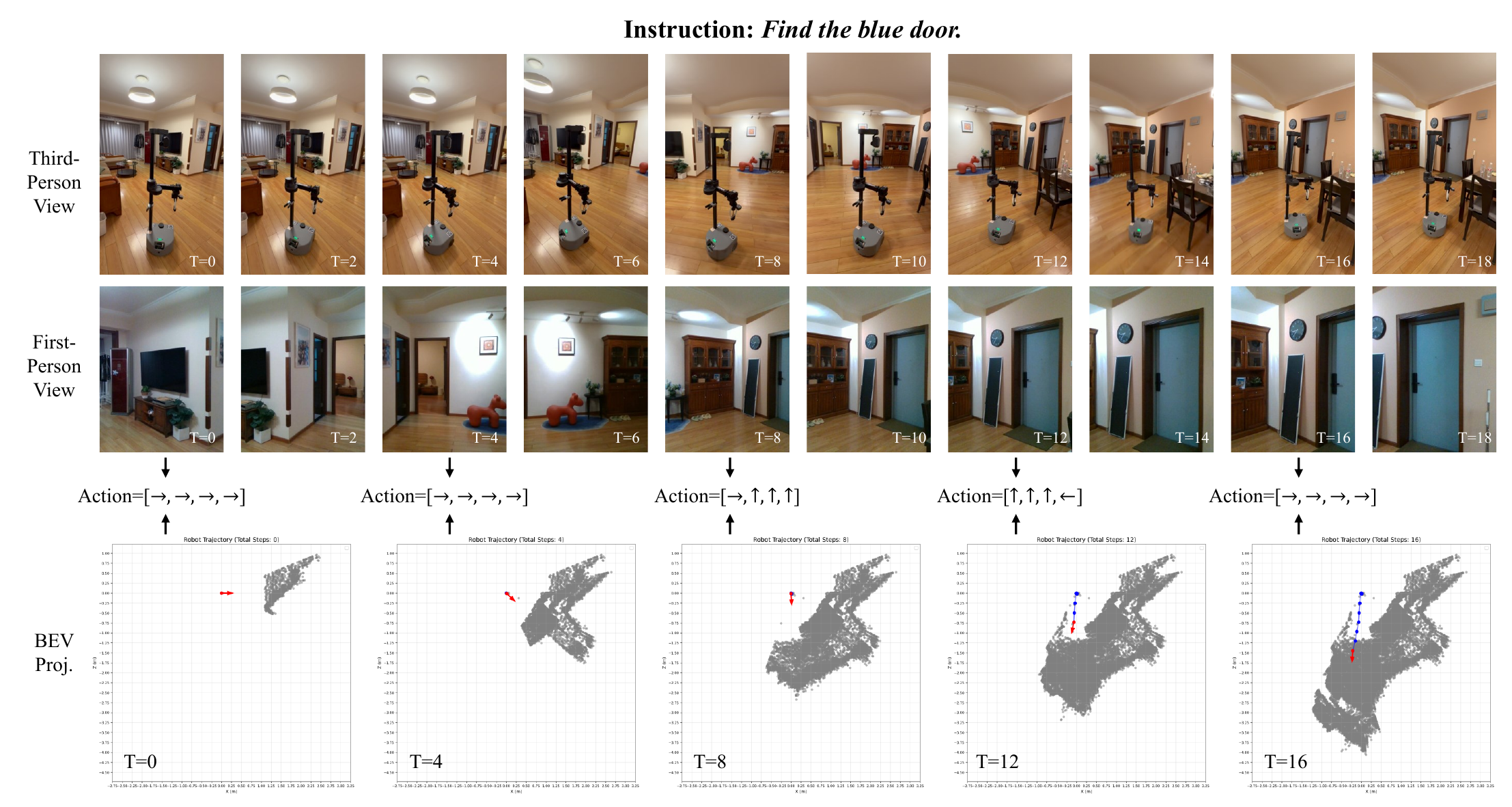}
    \caption{
        Real-world VLN results of GA-VLN: Example \#S1.
    }
    \label{fig:exmp2}
\end{figure*}

\begin{figure*}[b]
    \centering
    \includegraphics[width=1\textwidth]{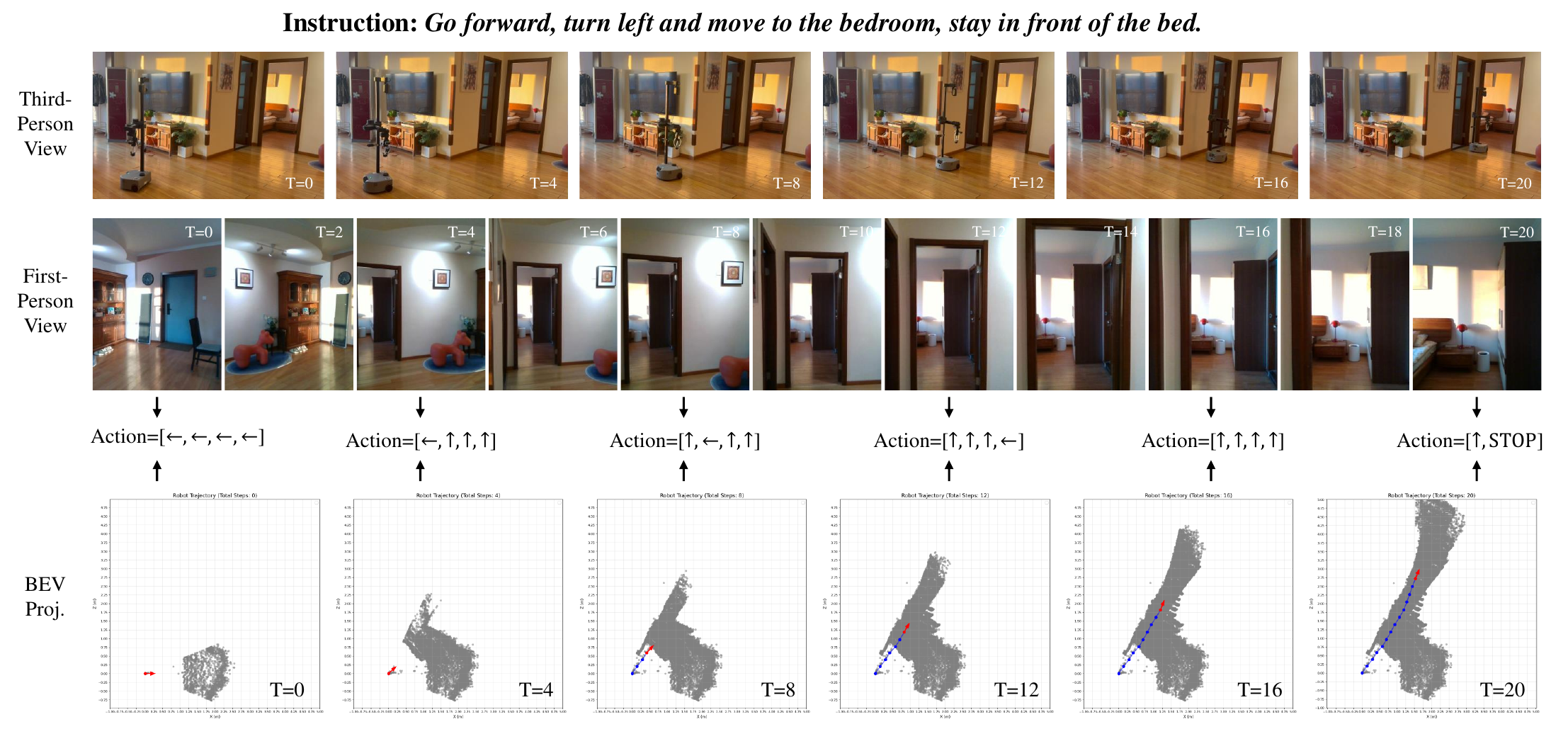}
    \caption{
        Real-world VLN results of GA-VLN: Example \#S2.
    }
    \label{fig:exmp3}
\end{figure*}

\paragraph{Setup.}
We use the Hello Robot Stretch 3 platform, a wheeled mobile system capable of executing low-level actions (e.g., moving forward and turning left/right) while providing RGB-D observations and odometry feedback. 
The agent's observations are transmitted via Wi-Fi to a local workstation for inference, after which the agent executes the predicted actions.
The experiments are performed in a real apartment of approximately 60\,m$^2$, closely matching the layout of the simulator environments. 
To mitigate the domain gap between simulation and reality, we introduced specific adaptations for stable deployment: (1) the rotational step angle was adjusted to $15^{\circ}$; (2) the RGB-D frame is captured at every step and the BEV representation aggregated up to 16 historical frames to handle real-world sensory noise; and (3) the two-round dialogue format described in Sec.~\ref{sec:action_predict} was disabled to streamline inference, the BEV representation is updated every four steps. All other settings remain identical to those used in the simulator-based experiments, and no additional modules (e.g., obstacle avoidance or navigable-point filtering) are introduced.

\begin{table*}[ht]
\small
\tabcolsep=0.1cm
\centering
\begin{tabular}{c |cccc|cccc|cccc}
\toprule
\multirow{2}{*}{Training Datasets}  
& \multicolumn{4}{c|}{R2R-CE} 
& \multicolumn{4}{c|}{RxR-CE} 
& \multicolumn{4}{c}{NavRAG-CE} 
\\
\cmidrule{2-13}
& NE\textdownarrow{} & OSR\textuparrow{} & SR\textuparrow{} & SPL\textuparrow{}
& NE\textdownarrow{} & OSR\textuparrow{} & SR\textuparrow{} & SPL\textuparrow{}
& NE\textdownarrow{} & OSR\textuparrow{} & SR\textuparrow{} & SPL\textuparrow{}
\\
\midrule
\textcolor{black}{w/o NavRAG-CE} 
& 4.80 & \textbf{67.6} & \textbf{61.0} & \textbf{55.2} 
& \textbf{5.88} & \textbf{67.0} & \textbf{55.4} & 45.2
& \textbf{7.88} & 46.4 & \textbf{22.2} & \textbf{18.2}
\\
\textcolor{black}{with NavRAG-CE} 
& \textbf{4.78} & 63.2 & 57.9 & 53.9 
& 5.98 & 64.7 & 54.3 & \textbf{46.3}
& 8.38 & \textbf{47.9} & 20.1 & 16.2
\\
\bottomrule 
\end{tabular}
\vspace{-3pt}
\caption{Ablation on training data composition across R2R-CE, RxR-CE, and NavRAG-CE benchmarks.}
\label{tab:supp1}
\end{table*}

\paragraph{Results.}
As shown in Fig.~\ref{fig:exmp2} and~\ref{fig:exmp3}, GA-VLN enables the agent to navigate accurately in real-world environments based on natural-language instructions, including both complex fine-grained descriptions (example \#1 and \#3) and high-level goal-directed commands (example \#2). The figures illustrate the navigation process from both third-person and first-person views, along with a visualization of the projection of visual patch features into the BEV space during execution. Especially from the first-person view, the navigation trajectories appear reasonable and coherent, and in demonstrated cases the agent successfully reaches the described targets (e.g., the bed, the television, or the blue door). For ease of demonstration, the BEV visualization is shown in an absolute coordinate frame; in practice, GA-BEV operates by projecting features into an ego-centric BEV coordinate system. These visualizations reveal that the agent maintains a meaningful geometric understanding of the environment—for example, it roughly covers the regions it has observed and is able to delineate major structural contours such as walls. Together, these qualitative results demonstrate the effectiveness of GA-VLN in real-world deployment.

\paragraph{Limitations.}
We observed distinct challenges in zero-shot real-world transfer. Without auxiliary obstacle avoidance modules, the agent occasionally executed paths dangerously close to obstacles (e.g., hugging walls), as it optimized for the shortest path trained in simulation. Additionally, the coarse granularity of discrete actions sometimes led to imprecise stopping behavior. These observations highlight the limitations of directly deploying the model in unmodified real-world environments. Nevertheless, despite operating without any auxiliary modules, GA-VLN shows strong instruction comprehension and produces reliable action sequences, indicating its potential as a foundational model for real-world navigation.

\section{More details of Geometry-Aware VLN Framework}
By integrating the proposed GA-BEV representation, we develop an efficient Vision-Language Navigation (VLN) framework that enables spatially grounded reasoning with compact visual tokens for navigation.

Specifically, at each navigation step $t$, given the language instruction $L$, the current-view visual features $V_t$, and the geometry-aware BEV features $B$ aggregated from $V_t$ and up to the last eight front-view observations $\{V_{t-1}, \dots, V_{t-8}\}$, the MLLM predicts an action sequence $A_{t}$ consisting of four discrete actions from the action vocabulary $\mathcal{A} = \{\textit{↑, ←, →,}\texttt{STOP}\}$:
\begin{equation}
A_{t} = f_{\text{MLLM}}(L, B, V_t)
\label{dia_1}
\end{equation}

After the agent completes the four actions in $A_t$, we denote its current front-view feature as $V_{t+1}$, and then proceed to predict $A_{t+1}$. Besides, we adopt a two-round dialogue format for action prediction, following~\cite{streamvln_wei2025streamvln}. Specifically, after the agent finishes the four actions in $A_t$, we do not immediately update the BEV feature $B$. Instead, the next prediction is made based only on the current $V_{t+1}$, $A_t$, and the input used in the previous prediction:
\begin{equation}
A_{t+1} = f_{\text{MLLM}}(L, B, V_t, A_t, V_{t+1})
\label{dia_2}
\end{equation}

After executing the four actions in $A_{t+1}$, the two-round dialogue process, Eq.~\ref{dia_1} and Eq.~\ref{dia_2}, is completed and a new dialogue begins. At this point, the current front-view feature becomes $V_{t+2}$. We then construct a new BEV feature using $V_{t+2}$ and up to eight previous frames $\{V_{t+1}, V_t, \dots, V_{t-6}\}$, and predict $A_{t+2}$ following Eq.~\ref{dia_1}. Therefore, the BEV feature is updated once every 8 actions.

During training, we divide each full trajectory into examples, each consisting of 8 actions, and train the model using the prompt format described above. During evaluation, we follow the two-round dialogue procedure, and the agent stops once it predicts the $\texttt{STOP}$ action.

\section{Additional Simulator Experiments}
\label{sec:additional_exp}

\paragraph{Ablation on Training Data Composition.}
Table~\ref{tab:supp1} reveals a trade-off in data composition. Incorporating NavRAG-CE data for making in-domain performance comparable to the finetuned model but adversely affected generalization on R2R-CE and RxR-CE benchmarks. This performance drop is likely attributed to the distributional shift in instruction styles across datasets. Consequently, to ensure robust generalization, we excluded NavRAG-CE from the primary pre-training data, only utilize it for fine-tuning.

\paragraph{Effect of Fusion Strategy.}
Table~\ref{tab:fusion} investigates the feature fusion strategy used in the our GA-BEV representation. 
Specifically, we compare two modes for fusing features from explicit depth-projected tokens and implicit 3D geometry tokens within the same BEV grid cell. 
Global mean pooling treats all tokens equally and computes their mean directly across sources, 
while hierarchical mean pooling first averages features of each modality separately and then fuses the two averaged representations. 
Results show that the global mean pooling achieves consistently better performance. 
We attribute this improvement to the higher resolution and richer patch-level representations of the 3D foundation model, whose pretrained geometric priors are better preserved under global fusion.
We attribute this improvement to its ability to preserve the fine-grained local feature distribution and maintain a balanced contribution between explicit and implicit geometry cues. 
In contrast, hierarchical fusion tends to oversmooth each modality before integration, weakening local geometric variations critical for accurate navigation.

\paragraph{Effect of BEV Update Step.}
As described in Sec.~\ref{sec:action_predict}, we adopt a structured navigation prompt for action prediction. 
The number of prompt-conversion rounds determines how frequently the BEV representation is updated during navigation. 
Table~\ref{tab:dataset_update_step} analyzes the effect of this BEV update interval on navigation performance. 
This parameter jointly influences both the spatial accuracy of the BEV representation and the number of training samples. 
Shorter BEV update intervals yield more temporally precise representations but substantially increase training cost due to finer trajectory segmentation.
Conversely, longer intervals reduce computation but lead to outdated spatial representations and weaker navigation performance.
Table~\ref{tab:dataset_update_step} shows that updating the BEV every 8 steps achieves the best balance between efficiency and accuracy, reducing total training time by nearly half with minimal performance loss compared with 4 updating steps.

\begin{table}[t]
  \centering
  \small
  \setlength{\tabcolsep}{6pt}
  \begin{tabular}{c|c c c c}
    \toprule
    \textbf{Fusion Strategy} 
    & \textbf{NE}\textdownarrow{} 
    & \textbf{OSR}\textuparrow{} 
    & \textbf{SR}\textuparrow{} 
    & \textbf{SPL}\textuparrow{}
    \\
    \midrule
    Global Mean Pooling & \textbf{5.03} & \textbf{59.60} & \textbf{53.56} & \textbf{49.41} \\
    Hierarchical Mean Pooling & 5.33 & 56.82 & 50.57 & 47.40 \\
    \bottomrule 
  \end{tabular}
  \caption{Comparison of BEV feature fusion strategies.}
  \label{tab:fusion}
\end{table}

\begin{table}[t]
  \centering
  \small
  \setlength{\tabcolsep}{6pt}
  \begin{tabular}{c |c c c c}
    \toprule
    \textbf{BEV Update Steps} 
    & \textbf{NE}\textdownarrow{} 
    & \textbf{OSR}\textuparrow{} 
    & \textbf{SR}\textuparrow{} 
    & \textbf{SPL}\textuparrow{}
    \\
    \midrule
    4 & 5.37 & \textbf{59.92} & \textbf{51.98} & 47.18 \\
    8 & \textbf{5.33} & 56.33 & 51.50 & \textbf{48.25} \\
    12 & 5.79 & 57.48 & 49.37 & 45.15 \\
    16 & 6.15 & 54.49 & 46.11 & 41.77 \\
    \bottomrule 
  \end{tabular}
  \caption{Ablation on BEV update interval w/o 3D geometry priors.}
  \label{tab:dataset_update_step}
\end{table}

\begin{table}[t]
  \centering
  \small
  \setlength{\tabcolsep}{6pt}
  \begin{tabular}{c|c| c c c c}
    \toprule
    \textbf{Setting} 
    & \textbf{Token Num} 
    & \textbf{NE}\textdownarrow{} 
    & \textbf{OSR}\textuparrow{} 
    & \textbf{SR}\textuparrow{} 
    & \textbf{SPL}\textuparrow{}
    \\
    \midrule
    RGB-Only & 4003 & 6.08 & 54.59 & 46.49 & 42.36 \\
    RGB-Depth & 8006 & 6.32 & 43.61 & 38.61 & 35.97 \\
    BEV Rep. & 394 & \textbf{5.33} & \textbf{56.33} & \textbf{51.50} & \textbf{48.25} \\
    \bottomrule 
  \end{tabular}
  \caption{Ablation on different depth processing strategies.}
  \label{tab:supp2}
\end{table}

\begin{figure*}[t]
\centering
\includegraphics[width=1\textwidth]{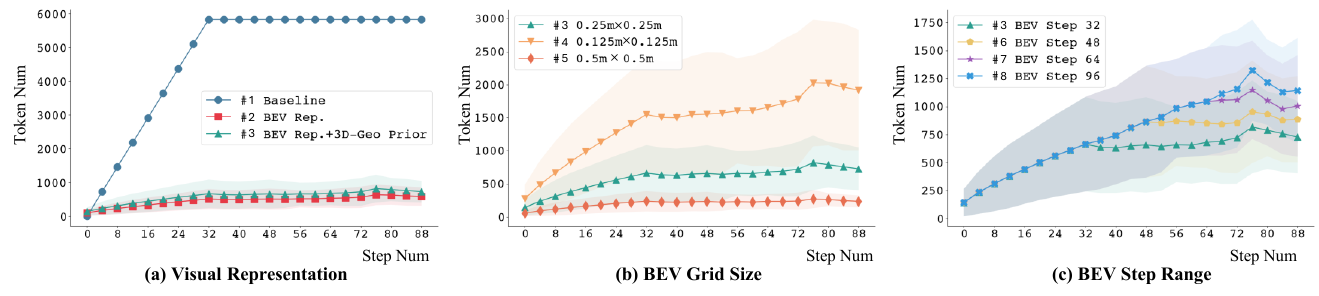}
\caption{
    Comparison of token usage across navigation steps for different configurations. The number shows in each legend corresponds to the configuration of the respective row in Table \ref{tab:tokennum}. The shaded area in the figure indicates the variance range of the sample data.
}
\label{fig:tokennum}
\end{figure*}

\paragraph{Ablation on Depth Processing Strategies.}
Unlike prior VLN methods~\cite{Navid_zhang2024navid, Uninavid_zhang2024uni, Navila_cheng2024navila, streamvln_wei2025streamvln} that rely solely on RGB observations, our approach additionally incorporates depth information. Table~\ref{tab:supp2} examines different strategies for processing depth and demonstrates the effectiveness and efficiency of our GA-BEV representation. The second row corresponds to a naive fusion strategy, where each RGB image is concatenated with a depth image encoded by the same visual encoder. This doubling of input tokens leads to a substantial increase in sequence length, imposing heavier computation while yielding noticeably worse performance across all metrics. In contrast, our BEV representation compresses both RGB and depth cues into a compact set of fewer tokens while achieving better performance. These results highlight that simply appending depth features is not an effective way to exploit geometric cues, whereas GA-BEV provides a more structured, geometry-aware representation that is both more accurate and significantly more token-efficient.

\begin{table}[t]
  \centering
  \small
  \setlength{\tabcolsep}{6pt}
  \begin{tabular}{c|c c c c}
    \toprule
    \textbf{Nav-VQA} 
    & \textbf{NE}\textdownarrow{} 
    & \textbf{OSR}\textuparrow{} 
    & \textbf{SR}\textuparrow{} 
    & \textbf{SPL}\textuparrow{}
    \\
    \midrule
    $\times$ & 4.80 & \textbf{67.59} & \textbf{60.96} & \textbf{55.19} \\
    \checkmark & \textbf{4.66} & 67.37 & 59.92 & 55.05 \\
    \bottomrule 
  \end{tabular}
  \caption{Ablation on the Nav-VQA Task.}
  \label{tab:supp3}
\end{table}

\paragraph{Experiments on the VQA Task.}
Compared with prior methods~\cite{Navila_cheng2024navila, streamvln_wei2025streamvln} that rely on large-scale general-purpose VQA datasets for co-training, our model does not incorporate any additional non-navigation data in order to maintain reasonable computational cost. Nevertheless, we conduct an auxiliary Nav-VQA experiment using only the navigation datasets. Following ~\cite{Navid_zhang2024navid, Navila_cheng2024navila}, the prompt is formatted as: \textit{``User: Assume you are a robot designed for navigation. You are provided with captured image sequences \texttt{<Images>}. Based on this image sequence, please describe the navigation trajectory of the robot. Assistant: \texttt{<instruction>}''}, where \texttt{<Images>} denotes 8 sampled RGB frames along the trajectory and \texttt{<instruction>} is the navigation instruction. As shown in Table~\ref{tab:supp3}, incorporating Nav-VQA supervision yields only marginal changes across all evaluation metrics. This suggests that our navigation datasets are sufficiently large and of high quality, enabling the MLLM to develop robust multimodal reasoning capabilities without requiring additional VQA-style data.

\paragraph{Visual Token Efficiency Across Navigation Steps.} 
To further illustrate the computational efficiency of our proposed GA-BEV representation, Figure~\ref{fig:tokennum} visualizes the accumulation of visual tokens across navigation steps for the various configurations detailed in Table~\ref{tab:tokennum}. As the agent explores the environment, the standard image-based baseline (Configuration \#1) exhibits a rapid, near-linear growth in token usage, quickly leading to an overwhelming computational burden and risking exceeding the MLLM's context window. In stark contrast, our GA-VLN variants (e.g., Configurations \#2 and \#3) maintain a remarkably low and stable token footprint throughout the entire long-horizon navigation trajectory. The shaded regions, which indicate the variance range of the sample data, further demonstrate the robustness and stability of our BEV updating mechanism across different environments and episode lengths. Additionally, the token accumulation trends for other hyperparameter settings (i.e., varying BEV grid sizes and step ranges corresponding to Rows \#4 - \#8) are also provided in the figure for comprehensive reference. Ultimately, this visualization confirms that compressing historical observations into a compact, ego-centric BEV space effectively eradicates the visual token redundancy issue, thereby enabling efficient and sustainable spatial reasoning for MLLMs in continuous environments.



\end{document}